\documentclass[runningheads]{llncs}
\usepackage[T1]{fontenc}

\usepackage{graphicx}
\usepackage{amsmath}

\usepackage{float}
\usepackage{booktabs}
\usepackage{multirow}
\usepackage{xcolor}
\usepackage{tcolorbox}
\usepackage{array, booktabs, amsmath}
\usepackage{tabularx}  
\usepackage{hyperref}

\usepackage{graphicx}   
\usepackage{booktabs}   
\usepackage{tabularx}   
\usepackage{adjustbox}  
\usepackage{siunitx}    
\usepackage{caption}    

\begin{document}
\title{Harnessing Chain-of-Thought Metadata for Task Routing and Adversarial Prompt Detection}
\titlerunning{Chain of Thought}

\author{Ryan Marinelli\inst{1}\orcidID{0009-0001-7279-3156
} \and
Josef Pichlmeier\inst{2} \and
\\ Tamas Bisztray\inst{1}\orcidID{0000-0003-2626-3434}}

\authorrunning{Marinelli et al.}

\institute{
    University of Oslo, Oslo, Norway\\
    \url{https://www.mn.uio.no/ifi/english/research/groups/sec/index.html}\\
    \email{ryanma@ifi.uio.no}
    \and
    BMW Group, Ludwig-Maximilians-Universit\"{a}t, Munich, Germany\\
    \email{Josef.Pichlmeier@bmw.de}
}

\maketitle           
\begin{abstract}
In this work, we propose a metric called ``Number of Thoughts (NofT)'' to determine the difficulty of tasks pre-prompting and support Large Language Models (LLMs) in production contexts. By setting thresholds based on the number of thoughts, this metric can discern the difficulty of prompts and support more effective prompt routing. A 2\% decrease in latency is achieved when routing prompts from the MathInstruct dataset through quantized, distilled versions of Deepseek with 1.7 billion, 7 billion, and 14 billion parameters. Moreover, this metric can be used to detect adversarial prompts used in prompt injection attacks with high efficacy. 
The Number of Thoughts can inform a classifier that achieves 95\% accuracy in adversarial prompt detection.  
Our experiments ad datasets used are available on our GitHub page: \url{https://github.com/rymarinelli/Number_Of_Thoughts/tree/main}.

\end{abstract}

\section{Introduction} Large Language Models (LLMs) have recently emerged as powerful tools capable of supporting significant advances in complex tasks such as software development and automated theorem proving, demonstrating impressive problem-solving capabilities despite lacking explicit symbolic reasoning mechanisms~\cite{FERRAG2025,bommasani2021opportunities,openai2025o3mini}.
At the core of the success of these models is the attention mechanism~\cite{10.5555/3295222.3295349}, enabling them to contextually generate coherent outputs beyond the capabilities of traditional sequential models, such as recurrent neural networks (RNNs). A widely adopted strategy to enhance this capability is the Chain-of-Thought (CoT) prompting technique~\cite{10.5555/3600270.3602070}, which encourages LLMs to explicitly generate intermediate problem-solving steps before arriving at a final answer. Although LLMs do not incorporate an explicit symbolic reasoning module, this step-by-step method—when properly prompted or fine-tuned—consistently improves performance on multi-step tasks.
Notably, OpenAI has recently indicated a strategic move toward incorporating CoT-like processes into the architecture rather than relying solely on prompting~\cite{openai2025gpt45}. This shift underscores the importance of exploring methods for leveraging intermediate step data—whether to route computations more efficiently or to detect adversarial logic—before such capabilities become natively embedded and potentially opaque to end users. 

When a CoT model is approaching a problem, it will segment the task into a number of intermediate steps to manage its complexity. Consequently, our underlying hypothesis is that the number of steps produced will  reflect the difficulty of the task, and also indicate the presence of deceptive logic, as observed in adversarial prompting scenarios. Our work centers around three research questions to explore the utilization of CoT-derived meta-information:
\begin{itemize}
    \item \textbf{RQ1:} Can the number of reasoning steps (``number of thoughts'') generated during Chain-of-Thought prompting serve as a reliable metric for estimating task complexity?
    
    \item \textbf{RQ2:} Can an abnormal or unexpected number of reasoning steps effectively indicate adversarial intent, such as prompt injection attacks?
    
    \item \textbf{RQ3:} Can the complexity estimation derived from the "number of thoughts" metric be effectively utilized for intelligent routing to optimize accuracy, latency, and power consumption?
\end{itemize}

In this work, we make the following key contributions:

\begin{enumerate}
    \item \textbf{Task Complexity Metric.} We propose and validate a novel metric called Number of Thoughts (NofT) for quantifying task complexity based on the number of intermediate reasoning steps generated in the Chain-of-Thought logic. We derive this metric by using a distilled Deepseek model combined with parsing logic to reliably infer the number of reasoning steps within the model's thought chain.

    \item \textbf{Complexity and Adversarial Intent Classifier.} We train and evaluate a robust Random Forest (RF) classifier capable of predicting the complexity of unseen prompts by estimating the number of required reasoning steps. Our results demonstrate that deviations from predicted complexity effectively indicate adversarial prompting attempts, such as prompt injection attacks, distinguishing these from naturally complex prompts.

    \item \textbf{Routing Optimization Framework.} We develop and evaluate a routing mechanism informed by our complexity metric, enabling efficient assignment of prompts to appropriately sized models. Experimental results show significant improvements in balancing accuracy, latency, and power consumption, demonstrating substantial operational cost savings and enhanced user experience.
\end{enumerate}

The rest of the paper is structure as follows. Section~\ref{sec:literature} overviews related work, Section \ref{sec:metric} introduces the NofT score and how it is calculated, while Section \ref{sec:class} discusses task difficulty estimation. Section \ref{sec:adv} presents how we perform adversarial prompt detection, and Section~\ref{sec:rout} the Routing Optimization Framework. Section \ref{sec:limit} discusses limitations while Section \ref{sec:conclusion} summarizes the results and concludes our work.

\section{Related Work} \label{sec:literature}

\subsection{The Emergence of Chain-of-Thought}

The original CoT prompting paper~\cite{10.5555/3600270.3602070} showed that providing step-by-step reasoning examples can dramatically improve LLM performance on complex tasks. Subsequent work found that even without exemplars, simply appending a cue like “Let’s think step by step” can elicit reasoning. This zero-shot CoT approach enabled large models to solve arithmetic and logic problems far better than standard zero-shot prompting (e.g. boosting accuracy on math tasks from ~10–20\% to 40\%+ on GSM8K)~\cite{NEURIPS2022_8bb0d291}.
Researchers at Google also improved how CoT answers are generated. Instead of producing one chain-of-thought, the model samples multiple diverse reasoning paths and then selects the most consistent final answer among them~\cite{52081}. The intuition is that complex problems may be solved via different reasoning routes, so aggregating multiple chains yields a higher chance of hitting a correct line of reasoning.

To tackle even harder tasks, various prompting strategies emerged utilizing basic CoT. Least-to-most prompting breaks a complex problem into a series of simpler sub-questions, solving them in sequence~\cite{zhou2022least}, while variants like ReAct combined CoT reasoning with actions (e.g. issuing tool use commands) and showed that interleaving thought steps with external actions can outperform reasoning-only approaches on interactive tasks~\cite{yao2023react}. 

With CoT gaining increasing popularity, researchers aimed to examine what factors could impact it's efficacy. 
One empirical finding is that the length of the reasoning chain matters. Jin et al. (2024) showed that longer reasoning sequences (even if they don’t add new information) can significantly boost performance, whereas overly short rationales hurt it~\cite{jin-etal-2024-impact}. Standard CoT imposes a latency cost (since the model must output long explanations). To mitigate this, compressed CoT methods encode reasoning internally. For example, Cheng et al. (2023) propose generating special “contemplation tokens” – dense latent representations that capture the content of a reasoning chain without producing the full text~\cite{cheng2024compressed}.
While initially Cot was simply a prompting technique, newer research it into model architectures annd training approaches. One direction is distillation of CoT abilities into smaller models by fine-tuning the latter on CoT datasets, thereby transferring multi-step reasoning skills~\cite{chen2025unveiling}.

The DeepSeek-R1 model is specifically engineered to autonomously generate Chain-of-Thought (CoT) reasoning without relying on external prompt templates or extensive prompting~\cite{deepseekR1}. Instead of explicitly instructed step-by-step reasoning, the model learns internally to structure and produce coherent reasoning chains. DeepSeek-R1 employs a structured, multi-stage training pipeline, beginning with an initial "cold start" phase, where it is fine-tuned on carefully curated high-quality, long-form CoT examples. Subsequently, the model undergoes a reasoning-focused reinforcement learning (RL) stage using Group Relative Policy Optimization (GRPO) to enhance reasoning capabilities further. The process concludes with additional refinement via rejection sampling and supervised fine-tuning, ensuring both performance and output readability.

\subsection{Routing Strategies in LLMs for Efficient Inference}

As LLM deployment grows, a key question is how to handle queries adaptively – routing each request to the best-suited model to optimize performance and cost. The two main directions for this are pre-generation assessment, or post-generation. For pre-generation a common approach is to train a router that predicts the complexity or difficulty of each input and chooses between a cheap model and an expensive model accordingly.
Ding et al. introduced Hybrid LLM, utilizing lightweight classifiers for difficulty estimation to effectively reduce expensive model calls by approximately 40\%, preserving overall accuracy~\cite{ding2024hybrid}. Similarly, RouteLLM employed learned routing policies to balance between strong and weak LLMs, achieving substantial cost reductions without sacrificing quality~\cite{ong_routellm_2024}.

Shnitzer et al. shows an approach for routing to the best model from a set of LLMs for a given task~\cite{shnitzer_large_2023} based on the HELM dataset~\cite{liang2022holistic}. TensorOpera Router enhances query efficiency, and leads to significant cost reductions by directing queries to domain-specific expert models~\cite{stripelis-etal-2024-tensoropera}. Other routing frameworks look at constraints like privacy to decide if a query should be processed on device or can be pushed to the cloud~\cite{zhang2024llm}. Collectively, these routing strategies demonstrate substantial computational cost savings and latency improvements, significantly enhancing LLM deployment efficiency and scalability.

\subsection{Metrics for Task Complexity and Routing Efficiency}

For a given task, a number of things  can be calculated before a prompt is routed including; the domain, required tools, token requirements, or the complexity of the task. We will solely focus on pre-generation assessment, as our own approach is targeting exactly that. 

Complexity metrics (predicted difficulty scores, query category labels) and cost metrics (fraction of big-model calls, latency saved) are used to characterize how well a routing approach adapts to input difficulty~\cite{ding2024hybrid,zhao2021calibrate}. Hybrid LLM reports the percentage of queries that invoked the expensive model, and RouteLLM quantifies overall cost reduction for a target accuracy.
However, these are mostly derived from the improvement of routing and are thus belong to the post-generation category.

Traditionally, if an LLM produced a chain-of-thought, we only evaluated whether the final answer was correct. However, a growing point of emphasis is evaluating the correctness and usefulness of the reasoning steps themselves. REVEAL, a benchmark dataset with human annotations for every step of a model-generated reasoning chain, each step in the chain is labeled for things like: relevance, attribution, and logical correctness~\cite{jin-etal-2024-impact}. In another study Nguyen et al. used knowledge graphs to assess CoT correctness. 

An insightful study by Tihanyi et al. showed that general models are surprisingly bad at estimating task difficulty~\cite{10825051}. In their study they prompted models with task, where the prompt heavily emphasized that the answer should be skipped if the model is not confident the answer could be correct. The only model from the study that showed signs of self reflection was OpenAI's o1-mini. This highlights the crucial importance of proper routing, as models like GPT-4o clearly lack the ability decide when a prompt should not be answered, or the answer should be skipped to avoid misleading users.

When it comes to adversarial prompting, Rossi et al. provides an overview of the current landscape of prompt injection and characterize the ongoing cat-and-mouse dynamic between developing safeguards for LLMs and the strategies attackers use to bypass them \cite{rossi2024earlycategorizationpromptinjection}. 
To the best of our knowledge, as of today nobody aimed train classifiers based on the number of steps in CoT processes to derive task difficulty, or to detect the presence of malicious prompts.

\section{Methodology for Deriving NofT} \label{sec:metric}
To summarize, the first important point to address in our study is to accurately measure and predict the number of reasoning steps that can be associated with a given prompt, since this measurement guides our the decisions behind routing and adversarial prompt detection. Section~\ref{sec:NofT}-Section~\ref{sec: RF} describes how we derive the number of thoughts (NofT) score, and train a Random Forest classifier for this purpose. Next, we take two datasets that have adversarial and non-adversarial prompt examples, and using this RF classifier we add an extra label to each prompt, namely the NofT score using the RF-1 classifier as explained in Section~\ref{sec:adv}. Then we divide these datasets to training and testing, and train two more RF classifiers on each dataset.
For the routing problem, we define three model sizes (small, medium, large) and show a strategy on how prompts can be routed by examining the NofT score of each prompt which is discussed in~\ref{sec:rout}.

\subsection{Deriving Number of Thoughts from a Training Dataset} \label{sec:NofT}
To our knowledge, no existing dataset uses the number of steps in the CoT process as an existing label.
To address this gap, we utilize the instruction-tuning dataset MathInstruct, which is compiled from 13 math datasets, 6 of which have CoT rationales~\cite{yue2023mammothbuildingmathgeneralist} and derive an estimate of the thought steps involved.

We select a subset of the dataset and employ a distilled version of Deepseek R1 \cite{deepseekR1} namely, DeepSeek-R1-Distill-Qwen-1.5B, to estimate the required number of reasoning steps. 
In the following, we detail how the number of CoT steps is derived:
To determine the number of CoT steps, we use a two-stage sentence analysis approach. First, we check if the model explicitly outputs the number of thoughts in the response. If this is missing, we fall back on a keyword based strategy. Here we count transitional words such as first, next, finally. These words are interpreted to indicate a shift in logic. This sentence analysis approach is applied twice per prompt, to derive an average number of detected steps to reduce variance. 

\begin{table}[htb]
\scriptsize
\centering
\caption{Parsed Response from DeepSeek-R1-Distill-Qwen-1.5B}
\begin{tabular}{@{}ll@{}}
\toprule
\textbf{Field} & \textbf{Content} \\ \midrule
\texttt{Question} & How many ways can the letters in the word \texttt{COMMON} be arranged? \\[1ex]
\texttt{Invoked Prompt} & Please provide a detailed, step-by-step solution to the following problem.\\
                 & Number each step sequentially. \\[1ex]
\texttt{Solution Steps} & 
\begin{tabular}[t]{@{}l@{}}
Step 1: Identify the letters in the word.\\[0.5ex]
Step 2: Determine the frequency of each letter.\\[0.5ex]
Step 3: Calculate the total number of arrangements.\\[0.5ex]
Step 4: Subtract the number of arrangements where two letters are swapped.\\[0.5ex]
Step 5: Sum the total arrangements and the subtracted arrangements.\\[0.5ex]
Step 6: Divide the total number of arrangements by the number of duplicate letters.
\end{tabular} \\
\label{tab: example_parse}
\texttt{Thought Count} & 6 \\
\bottomrule
\end{tabular}
\end{table}

Table \ref{tab: example_parse} shows our methodology for deriving the number of steps for each task in the benchmark. Each prompt sent to the model consists of the question, and an invoked prompt that stays the same for every query. In the latter the model is instructed to number each step of their CoT process, as shown under ``Solution Steps''.


\subsection{Estimating the Number of Thoughts}
\label{sec: RF}
For both the routing and malicious prompt detection experiments, a fast and robust method for predicting the number of thought steps is required. In our study, we train a Random Forest (RF) classifier on the MathInstruct dataset with the labels being the number of reasoning steps (Thought Count in Table~\ref{tab: example_parse}. 
RF works by using an ensemble of decision trees with each trained on a random sample of the data. The final prediction is derived by averaging the outputs of all the trees, which helps to reduce overfitting and improve overall performance \cite{breiman2001random}.

As we want to derive the number of thoughts based on the query, it is necessary to first derive a valid representation of the prompts. For this, the Term Frequency-Inverse Document Frequency (TF-IDF) vectorization is used to convert the questions into a numeric representation. Thereby, the Term Frequency is simply the ratio of the number of times a word appears in prompt to the total number words in the prompt. The Inverse Document Frequency is the log ratio of the total number of prompts to the number of prompts that contain a keyword. Multiplying these values gives the TF-IDF score which captures the uniqueness of a particular input relative to the dataset \cite{pedregosa2011scikit}. 

In contrast to contextual embeddings like those generated by BERT or BAAI BGE \cite{BERT_2018,BGE_2024}, TF-IDF vectorization gives a sparse representation that can be used for classifiers such as RF. Dense embeddings may not capture enough context when the number of samples within the dataset is limited. Furthermore, using embeddings from a specific model could constrain the estimator’s performance while introducing additional bias \cite{bolukbasi2016man}, especially given the domain specific math expressions used in this study.

\begin{equation}
\hat{y}(x) = \frac{1}{N} \sum_{i=1}^{N} T_i(x)
\end{equation}
where $T_i(x)$ is the prediction of the $i^\text{th}$ tree.

The model's objective is to minimize the discrepancy between the thought counts parsed from Deepseek and the predicted number of thoughts. For training the RF, the discrepancy is captured using the Mean-Squared Error of the classifier. 

\begin{equation}
\text{MSE} = \frac{1}{n} \sum_{j=1}^{n} \left( y_j - \hat{y}(x_j) \right)^2,
\end{equation}
where $n$ is the number of samples, $y_j$ is the thought count from Deepseek's response for $j^\text{th}$ prompt, and $\hat{y}(x_j)$ is the estimated number of thoughts using RF.
The trained RF algorithm allows us to estimate the number of thoughts that would be needed for an CoT LLM to solve a specific problem.

\subsection{Ablation Study of Model Size}
In order to further test our approach and investigate the impact of model scaling and estimation methods on the number of thoughts metric, we perform an ablation study. The analysis compares different configurations including quantized versions of the Deepseek R1 model ranging from 1.5B to 14B parameters. Furthermore, we introduce an external evaluator, namely a quantized version of the TinyR1 \cite{tinyr1proj} model. In addition the RF classifier presented in Section \ref{sec: RF} is used as the statistical baseline for estimating the number of reasoning steps. The RF classifier is tested by examining the average number of thoughts across different levels of query difficulty. Along that, we present t-test results, power analysis, and Bayesian simulations. 
Due to the heavy computational requirements we only use a subset of 20 prompts contained in Dynamic Intelligence Assessment Benchmark(DIA)~\cite{10825051} for the ablation study.

\subsection{Statistical Analysis of Model Estimation}

We evaluate the performance using two metrics. First, we calculate the average number of thoughts for prompts categorized as Easy, Medium, and Hard. Second, we perform t-tests to obtain p-values that compare the mean thought counts between these difficulty levels. The two metrics help us to determine if there is a statistical significant difference between the prompts of varying complexity; this is required to inform the classifier.
Table \ref{tab:summary} summarizes the results. The RF classifiers approximates the average thought count of the Deepseek models well. This is especially significant as the RF was trained on a different dataset, indicating that it can effectively capture the number of reasoning steps. Furthermore, this suggests that it can be effectively used for routing architectures. 

The TinyR1 model shows limitations in its prompting strategies. It returns large text blocks with few clear transition, causing the parsing logic to miss sub-steps bundled within larger steps. With respect to the difficulty level, the 14B model better distinguishes between easy and medium prompts. The smallest 1.5B model shows the following trend, easy prompts require fewer steps while hard prompts require more. 

\begin{table}[htb]
\centering
\fontsize{11pt}{13pt}\selectfont  
\renewcommand{\arraystretch}{1.3} 

\captionsetup{skip=5pt}  

\resizebox{\linewidth}{!}{%
\begin{tabular}{ll S[table-format=2.2] S[table-format=2.2] S[table-format=2.2] | S[table-format=1.3] S[table-format=1.3] S[table-format=1.3]}
\toprule
\textbf{Dataset} & \textbf{Metric} 
  & \textbf{Easy} \hspace{10pt} 
  & \textbf{Medium} \hspace{10pt} 
  & \textbf{Hard} \hspace{15pt} 
  & $\mathbf{\bar{X}_{\text{Easy}} - \bar{X}_{\text{Med}}}$ \hspace{10pt} 
  & $\mathbf{\bar{X}_{\text{Easy}} - \bar{X}_{\text{Hard}}}$ \hspace{10pt} 
  & $\mathbf{\bar{X}_{\text{Med}} - \bar{X}_{\text{Hard}}}$ \\
\midrule
\textbf{TinyR1 Quantized (32B)} 
  & Thought Count  
  & 5.80 & 5.80 & 5.00 
  & \textbf{1.000} 
  & \textbf{0.390} 
  & \textbf{0.343} \\
\midrule
\multirow{5}{*}{\textbf{Deepseek Quantized}} 
  & Thought Count (1.5B) 
  & 8.80 & 9.10 & 11.90 
  & \textbf{0.959} 
  & \textbf{0.622} 
  & \textbf{0.603} \\
  & Thought Count (7B) 
  & 6.70 & 5.50 & 5.50  
  & \textbf{0.253}  
  & \textbf{0.663}  
  & \textbf{0.578} \\
  & Thought Count (14B) 
  & 5.70 & 7.30 & 7.20  
  & \textbf{0.036} 
  & \textbf{0.279} 
  & \textbf{0.945} \\
  & Average Thought Count 
  & 7.05 & 7.15 & 8.12  
  & \multicolumn{1}{c}{--} & \multicolumn{1}{c}{--} & \multicolumn{1}{c}{--} \\
  & Intergrader Variability (SD) 
  & 1.58 & 1.85 & 3.32  
  & \multicolumn{1}{c}{--} & \multicolumn{1}{c}{--} & \multicolumn{1}{c}{--} \\
\midrule
\textbf{Random Forest} 
  & Predicted Thought Count 
  & 7.58 & 7.47 & 8.06 
  & \textbf{0.064} 
  & \textbf{0.004} 
  & \textbf{0.001} \\
\bottomrule
\end{tabular}%
}
\caption{Summary statistics and p-values for differences of means in TinyR1, Deepseek Quantized, and RF.}
\label{tab:summary}
\end{table}

\subsubsection{Power Analysis}

Power analysis allows to determine if a statistical test will detect a true effect, i.e., correctly reject a false null hypothesis \cite{dorey2011statistics}. In the context of this study, it measures the likelihood that the differences in thought counts among easy, medium and hard prompts represents real effects rather than random noise. Power is influenced by the effect size measured using Cohen's D, sample size, the significance level often set to 0.05, and the variability in the data.

A key limitation arises from the DIA dataset. It only contains 20 prompts classified as hard. This limits the statistical power and requires caution interpretation of the p-values. 

\begin{table}[htb]
\centering
\fontsize{11pt}{13pt}\selectfont  
\renewcommand{\arraystretch}{1.3} 
\captionsetup{skip=5pt}  

\resizebox{\linewidth}{!}{%
\begin{tabular}{l|l|c|c|c|}
\hline
\textbf{Model} & \textbf{Comparison}  & \textbf{Effect Size (Cohen's d)}  & \textbf{Achieved Power}  & \textbf{Required Sample Size}  \\
\hline
\multirow{3}{*}{\textbf{TinyR1}}  
  & easy vs hard    & 0.000 & - & -  \\
\cline{2-5}
  & easy vs medium  & \textbf{0.816} & \textbf{0.710} & \textbf{24.60}  \\
\cline{2-5}
  & hard vs medium  & \textbf{0.969} & \textbf{0.847} & \textbf{17.74}  \\
\hline
\multirow{3}{*}{\textbf{Deepseek (1.5B)}}  
  & easy vs hard    & 0.498 & 0.335 & 64.36  \\
\cline{2-5}
  & easy vs medium  & 0.052 & 0.053 & 5754.59  \\
\cline{2-5}
  & hard vs medium  & 0.526 & 0.367 & 57.76  \\
\hline
\multirow{3}{*}{\textbf{Deepseek (7B)}}  
  & easy vs hard    & 0.440 & 0.273 & 82.24  \\
\cline{2-5}
  & \textbf{easy vs medium}  & \textbf{1.160} & \textbf{0.947} & \textbf{12.70}  \\
\cline{2-5}
  & hard vs medium  & 0.562 & 0.410 & 50.73  \\
\hline
\multirow{3}{*}{\textbf{Deepseek (14B)}}  
  & \textbf{easy vs hard}    & \textbf{1.109} & \textbf{0.927} & \textbf{13.79}  \\
\cline{2-5}
  & \textbf{easy vs medium}  & \textbf{2.193} & \textbf{1.000} & \textbf{4.46}  \\
\cline{2-5}
  & hard vs medium  & 0.069 & 0.055 & 3273.94  \\
\hline
\multirow{3}{*}{\textbf{Random Forest}}  
  & \textbf{easy vs hard}    & \textbf{2.122} & \textbf{1.000} & \textbf{4.67}  \\
\cline{2-5}
  & easy vs medium  & 0.379 & 0.215 & 110.24  \\
\cline{2-5}
  & \textbf{hard vs medium}  & \textbf{2.111} & \textbf{1.000} & \textbf{4.71}  \\
\hline
\end{tabular}%
} 

\caption{Power Analysis results for TinyR1, Deepseek (1.5B, 7B, 14B), and RF models. Bold values indicate significant power.}
\label{tab:power_analysis}
\end{table}

For TinyR1, the effect size was large enough that the achieved power was .84 for the difference between hard and medium. .8 is a general threshold for being sufficient. Thus, TinyR1 cannot be used distinguish between hard and medium prompts.

For the Deepseek models, there generally was not enough power. The power achieved for the 14B model for the difference of means in easy and medium prompts was 1 with the p-value being less than .05. Thus, the can confidently segregate between easy and medium questions. There is also enough power to determine that 14B cannot distinguish between easy and hard prompts with the power being .92 and the p-value being .279. 

The RF classifier has enough power to determine easy and hard prompts with a power of 1. This is also the case for hard and medium prompts. The effect size is to small for it to discern between easy and medium with a power of .21. The p-values for each difference of mean is well under .05. Given the power achieved, given the extreme values, one can determine the RF classifier can discern the different difficulties across prompts. 

\subsubsection{Bayesian Approach}
Given the lack of power, Bayesian simulation is also applied to asses the differences in number of thoughts across the query difficulty. We use the calculated means and standard deviations from the different models to form the prior, and we assume that the data for each difficulty is normally distributed. The standard error is calculated by dividing the standard deviation by the square root of the sample size. We then use Markov Chain Monte Carlo (MCMC) sampling with 2000 draws and a target acceptance of 0.9 to generate the posterior distribution for the mean thought counts.

The results for the TinyR1 model are shown in Table \ref{tab:tinyr1-bayes}. The posterior means are 5.795 for Easy, 4.977 for Medium, and 5.805 for Hard prompts. The difference between Easy and Hard is minimal and uncertain, while the difference between Medium and Hard is slightly more significant at -0.808. However, the Highest Density Interval (HDI) includes zero, making the evidence inconclusive.

\begin{table}[H]
\centering
\begin{tabular}{lrrrrrr}
\toprule
\(\textbf{Parameter}\) & \(\mathbf{\mu}\) & \(\mathbf{\sigma}\) & \(\text{HDI}_{3\%}\) & \(\text{HDI}_{97\%}\) & \(\text{MCSE}_{\mu}\) & \(\text{MCSE}_{\sigma}\) \\
\midrule
\(\mu_{\mathrm{easy}}\)            & 5.795 & 0.725 & 4.454 & 7.150 & 0.011 & 0.008 \\
\(\mu_{\mathrm{medium}}\)          & 4.997 & 0.637 & 3.812 & 6.194 & 0.009 & 0.006 \\
\(\mu_{\mathrm{hard}}\)            & 5.805 & 0.495 & 4.809 & 6.699 & 0.008 & 0.006 \\
\(\delta_{\mathrm{easy,\,hard}}\)   & -0.010 & 0.867 & -1.666 & 1.569 & 0.014 & 0.014 \\
\(\delta_{\mathrm{medium,\,hard}}\) & -0.808 & 0.805 & -2.300 & 0.674 & 0.012 & 0.010 \\
\bottomrule
\end{tabular}
\caption{Bayesian posterior summaries for TinyR1.}
\label{tab:tinyr1-bayes}
\end{table}

For Deepseek, the smallest model is used to form the prior. The posterior means are 8.677, 8.995, and 11.255 for easy, medium, and hard prompts respectively. These values suggest more signification differences between the difficulty levels compared to TinyR1. Yet, even here, the HDIs for the differences such as -2.5 for Easy vs. Hard include zero, so the evidence remains somewhat tentative.

\begin{table}[thb]
\centering
\begin{tabular}{lrrrrrr}
\toprule
\(\textbf{Parameter}\) & \(\mathbf{\mu}\) & \(\mathbf{\sigma}\) & \(\text{HDI}_{3\%}\) & \(\text{HDI}_{97\%}\) & \(\text{MCSE}_{\mu}\) & \(\text{MCSE}_{\sigma}\) \\
\midrule
\(\mu_{\mathrm{easy}}\)             & 8.667  & 3.133  & 2.854  & 14.456  & 0.051  & 0.036 \\
\(\mu_{\mathrm{medium}}\)           & 8.995  & 1.999  & 5.331  & 12.763  & 0.032  & 0.023 \\
\(\mu_{\mathrm{hard}}\)             & 11.255 & 3.345  & 4.885  & 17.311  & 0.053  & 0.038 \\
\(\delta_{\mathrm{easy,\,hard}}\)    & -2.588 & 4.599  & -11.365 & 5.859   & 0.073  & 0.063 \\
\(\delta_{\mathrm{medium,\,hard}}\)  & -2.260 & 3.879  & -9.501 & 4.944   & 0.063  & 0.053 \\
\bottomrule
\end{tabular}
\caption{Bayesian posterior summaries for Deepseek}
\label{tab:bayes-summaries}
\end{table}

For the RF classifier, the posterior means are 7.58, 7.47, and 8.063 for easy, medium, and hard prompts respectively. 
Importantly, the HDI for the difference between easy and hard prompts does not include zero, indicating a statistical significant difference. Same for the medium vs hard prompts. This suggests that the RF classifier can differentiate between the difficult levels on thought counts.

\begin{table}[thb]
\centering
\begin{tabular}{lrrrrrr}
\toprule
\(\textbf{Parameter}\) & \(\mathbf{\mu}\) & \(\mathbf{\sigma}\) & \(\text{HDI}_{3\%}\) & \(\text{HDI}_{97\%}\) & \(\text{MCSE}_{\mu}\) & \(\text{MCSE}_{\sigma}\) \\
\midrule
\(\mu_{\mathrm{easy}}\)            & 7.584 & 0.023 & 7.541 & 7.626 & 0.000 & 0.000 \\
\(\mu_{\mathrm{medium}}\)          & 7.471 & 0.055 & 7.366 & 7.569 & 0.001 & 0.001 \\
\(\mu_{\mathrm{hard}}\)            & 8.063 & 0.138 & 7.807 & 8.325 & 0.002 & 0.002 \\
\(\delta_{\mathrm{easy,\,hard}}\)   & -0.480 & 0.139 & -0.725 & -0.206 & 0.002 & 0.002 \\
\(\delta_{\mathrm{medium,\,hard}}\) & -0.592 & 0.147 & -0.866 & -0.309 & 0.002 & 0.002 \\
\bottomrule
\end{tabular}
\caption{Bayesian posterior summaries for RF}
\label{tab:tinyr1-bayes}
\end{table}

\subsubsection{Ablation Review}

As presented in the previous sections, we used a combination of conventional statistical techniques and Bayesian methods. Our goals was to evaluate how well our models distinguish between different prompt difficulty levels. To achieve this, we compared the mean number of thought steps across difficulty levels. The CoT models achieved mixed results in their ability to discern differences in the number of thoughts for the different levels. The RF classifier consistently and statistically significantly distinguishes between prompts and varying complexity.

\section{Prompt Difficulty Estimation} \label{sec:class}

We use the RF classifier for two tasks, assessing the prompt difficulty and detecting adversarial prompts. To achieve this, we leverage the DIA dataset~\cite{10825051}, which contains prompts annotated by experts. Each prompt is labeled as easy, medium, or hard, and includes a boolean flag for whether the prompt is adversarial. These annotations serve as our ground truth for evaluating the performance of classifiers in both gauging difficulty and identifying adversarial intent.

\subsection{Gauging Difficulty}
Gauging difficulty in this setting is essentially observing agreement between the model and expert annotations. 
Difficulty is a bit ambiguous to capture, so the ground truth is not as strongly defined. While the DIA is annotated by experts, difficulty is more of a qualitative construct that is attempting to be quantified. 

\begin{figure}
    \centering
    \includegraphics[width=1.0\linewidth]{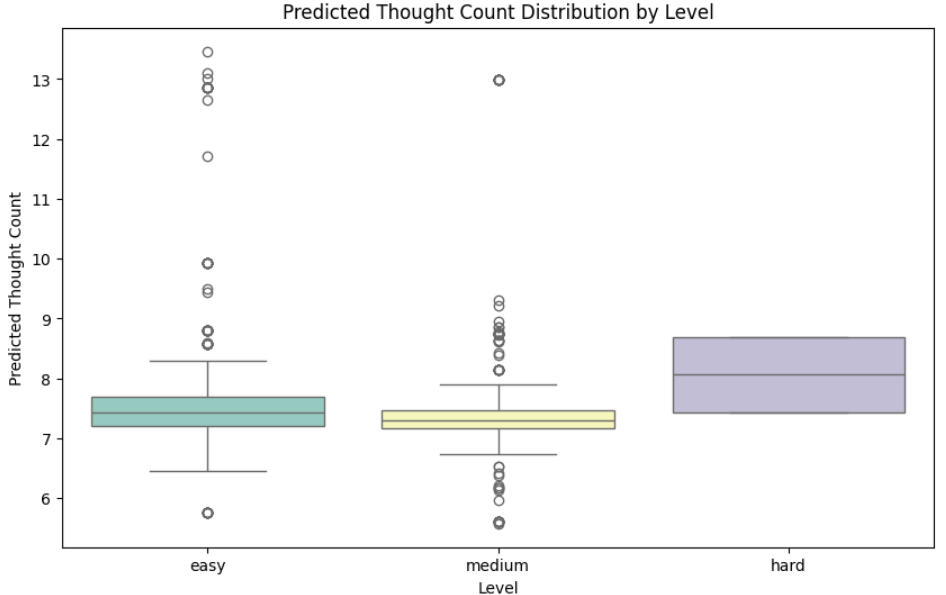}
    \caption{Predicted Thought Count and Prompt Difficulty}
    \label{fig:enter-label}
\end{figure}

When comparing the distributions of predicted number of thoughts and their overlap with the difficulty, there are noticeable differences. The mean number of thoughts for easy and hard show a clear distinction. The Medium category, as expected, is less well-defined.  Figure \ref{fig:enter-label} illustrates the relationship between predicted thought counts and prompt difficulty showcasing the mentioned differences.

\subsection{Imbalanced Learning}
Due to the class imbalance of the different difficulty levels, Synthetic Minority Oversampling Technique(SMOTE) is applied. In essence, SMOTE samples from the minority class, in our case the most difficult prompts, and generates additional synthetic examples based on the nearest neighbors. SMOTE augments the dataset and therefore improves the learned decision boundary.

\subsection{Hyperparameter Tuning}
Hyperparameters are configuration settings that determine how the RF model learns and performs during inference. This includes fine-tuning settings such as the number of trees, their depth the the minimum number of samples required to split a node, and the use of bootstrapping. In this study, we use a random search to systematically explore the hyperparameter space and identify the best configuration.

\subsection{Calibration}

Calibration adjusts a model's output to better match the true class frequencies. Models often produce probability estimates that are poorly calibrated which might result in the assignment of high probabilities to a minority class. Calibration ensures that a model's confidence levels correspond more closely to the actual occurrence rates of the classes in the data.

In this paper we use isotonic regression for calibration. The goal is to enforce that the predicted probabilities match the observed class frequencies. It achieves this by fitting a function that maps the raw probability estimates to calibrated probabilities, minimizing the error between predicted probabilites and class frequencies \cite{barlow1972statistical}.

\subsection{Prompt Difficulty Classification Results}
In order to interpret the results of the model, it is useful to establish common ground in terms of the metrics used in the evaluation. Precision is the proportion of true positives to the total of true positives and false positives. Recall is the proportion of true positives to the total of true positives and false negatives. Essentially, precision captures how many of the model's predictions as positive were true positive. Recall captures how many of the true positives were correctly identified by the model. The F1 Score is the harmonic mean of precision and recall; it captures a balance between both in a single metric. Support is the number of instances of each class in the dataset.

The classifier achieved an overall accuracy of 80\%. For prompts labeled easy, 98\% were correct. Medium prompts were predicted with 64\% accuracy while hard prompts where only classified correctly 35\% of the time.

The recall for hard prompts is much higher than the accuracy. For medium prompts it reaches 90\%. In the case of classifying hard prompts, it seems the model is able to determine most of the hard prompts, but it may misclassify medium prompts as hard prompts.

The F1 scores for Easy and Medium prompts were higher than those for Hard prompts. This discrepancy can be partially explained by the lower number of hard prompt instances in the dataset, which may contribute to increased variability in performance for that class.

\begin{table}[ht]
\centering
\begin{tabular}{lcccc}
\hline
\textbf{Class} & \textbf{Precision} & \textbf{Recall} & \textbf{F1-Score} & \textbf{Support} \\
\hline
easy   & 0.98 & 0.77 & 0.87 & 220 \\
medium & 0.64 & 0.90 & 0.75 & 73  \\
hard   & 0.25 & 0.86 & 0.39 & 7   \\
\hline
\multicolumn{5}{c}{\textbf{Overall Metrics}} \\
\hline
Accuracy      & \multicolumn{3}{c}{0.81} \\
\hline
\end{tabular}
\caption{DIA Prompt Difficulty Classification}
\label{tab:classification_report}
\end{table}

\section{Adversarial Prompt Detection} \label{sec:adv}
In order to assess whether the number of thoughts can inform prompt injection, we employed the prompt injection dataset from deepset \cite{deepset_prompt_injections}. We predicted the number of thoughts for each prompt and then performed an independent t-test comparing the means between adversarial and non-adversarial prompts. With a computed t-statistic of -4.19, which exceeds the critical threshold for $\alpha$ = 0.05, we reject the null hypothesis of no difference between the group means. This result indicates a statistically significant difference in the mean predicted number of thoughts between the two groups.

\subsection{Classification Results with Deepset Dataset}
Using a similar procedure as with the previous classifier, the model achieves a nearly equal amount of precision between classes. However, the recall for the non-adversarial prompts is about double than the adversarial. It is possible that the model is being conservative and making fewer false positives which is seen in the higher precision metric. However, that comes at the cost of missing some true positives. This may be more desirable behavior given the nature of the classifier. 

\begin{table}[ht]
\centering
\begin{tabular}{lcccc}
\toprule
Class & Precision & Recall & F1-Score & Support \\
\midrule
Non-Adversarial& 0.73 & 0.91 & 0.81 & 69 \\
Adversarial & 0.75 & 0.44 & 0.55 & 41 \\
\midrule
Accuracy & \multicolumn{4}{c}{0.74} \\
\bottomrule
\end{tabular}
\caption{Deepset Classification Metrics}
\label{tab:classification_report_before}
\end{table}

\subsection{Classification Results with DIA-Bench}
To determine the threshold for the classes, the ROC curve is used to inform on the trade-offs between precision and accuracy. The threshold determined is .90 meaning the model will label a prompt as adversarial, only if the probability of it being adversarial is 90\% or above. 

\begin{figure}[H]
    \centering
    \includegraphics[width=0.5\linewidth]{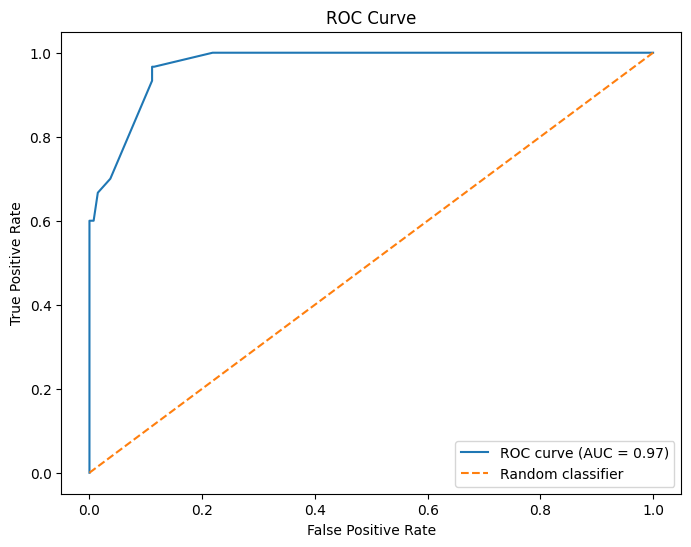}
    \caption{Determining Threshold}
    \label{fig:ROC_curve}
\end{figure}

As shown in Figure \ref{fig:ROC_curve}, the model does well in balancing precision and recall with this additional threshold consideration. It achieves 95\% accuracy. The main reason is that it must be more certain before labeling a prompt as adversarial. Essentially this is instilling domain expertise into the evaluation. Similar tuning was attempted to the deepset dataset, but it did not significantly improve results. 
\begin{table}[H]
\centering
\begin{tabular}{lcccc}
\toprule
Class & Precision & Recall & F1-Score & Support \\
\midrule
Not Adversarial & 0.96 & 0.99 & 0.97 & 270 \\
Adversarial  & 0.90 & 0.60 & 0.72 & 30 \\
\midrule
Accuracy & \multicolumn{4}{c}{0.95} \\
\bottomrule
\end{tabular}
\caption{DIA Classification Metrics}
\label{tab:final_classification_report}
\end{table}

\section{Routing} \label{sec:rout}

With the number of thoughts established as a reliable metric for measuring prompt difficulty, we extend its application to model routing. If a rather simple statistical model can infer query complexity, it can be used in a routing mechanism to distribute prompts to the most appropriate model. For example, easy prompts can be routed to smaller, more resource efficient models, while harder prompts are forwarded to larger, more capable models. To determine the optimal routing strategy, we consider multiple metrics: number of thoughts, latency, power consumption, and accuracy. Latency is measured in seconds, power consumption in joules, and accuracy employing the ROUGE score. 

\subsection{Models}
Three distilled versions of the Deepseek R1 model are compared at different scales in our routing mechanism. This includes versions with 1.5 billion, 7 billion, and 14 billion parameters. Due to resource constraints, we use INT8 quantized versions of these models. While this reduces model accuracy, it reduces the memory requirements significantly. Since all models are quantized, the relative performance comparison stays valid. 

\subsection{Metrics}

To calculate the energy consumption we are using following equation: 

\begin{equation}
E = P_{\text{avg}} \times t
\label{eq: energy}
\end{equation}

\noindent
where:
\begin{itemize}
    \item \(E\) is the energy consumed (in Joules),
    \item \(P_{\text{avg}}\) is the average power (in Watts), and
    \item \(t\) is the latency or time taken (in seconds).
\end{itemize}

Power consumption is measured by running a thread in the background and collecting the usage data through the NVIDIA Management Library (NVML). Latency is measured as the time elapsed from the start of the model inference until the complete response is received. The final energy consumption is calculated by averaging the measured power over the inference period and then multiplying it by the measured latency. 

\begin{equation}
t = t_{\text{end}} - t_{\text{start}}
\end{equation}

\noindent
where:
\begin{itemize}
    \item \(t\) is the total latency (in seconds),
    \item \(t_{\text{start}}\) is the starting time of response, and
    \item \(t_{\text{end}}\) is the ending time of response
\end{itemize}

Recall-Oriented Understudy for Gisting Evaluation (ROUGE) is used to determine the accuracy of the responses. While ROUGE can be expressed in several metrics (ROUGE-N, ROUGE-L, ROUGE-S) only ROUGE-L is used in this study. The metrics captures the longest common subsequence(LCS) between a generated text and a target and therefore is a robust indicator for content similarity. 

\begin{equation}
\text{ROUGE-L} = \frac{(1 + \beta^2) \, R_{LCS} \, P_{LCS}}{R_{LCS} + \beta^2 \, P_{LCS}}
\end{equation}

\noindent
where:
\begin{align*}
R_{LCS} &= \frac{\text{LCS}(\text{Target}, \text{Generated})}{\text{Length of Target}}, \\
P_{LCS} &= \frac{\text{LCS}(\text{Target}, \text{Generated})}{\text{Length of Candidate}},
\end{align*}
and \(\beta\) is a parameter to balance recall and precision. In our evaluation, $\beta$ is set to one to assign equal importance to both metrics.

Given that the models tend to produce verbose responses during their reasoning process, ROUGE-L offers a robust basis for comparing the generated output with the ground truth in the dataset.

\subsection{Threshold Determination}
To determine how prompts should be routed across models of different sizes, we calculate two thresholds. One threshold indicates the minimum thought count at which it is better to switch from the 1.5 billion-parameter model to the 7 billion-parameter model. A second threshold indicates the additional thought count needed to justify the 14 billion-parameter model over the 7 billion-parameter option.

\begin{align*}
\overline{R}\big(M(n_i)\big) &= \text{Average ROUGE-L score for the selected model}, \\
\overline{L}\big(M(n_i)\big) &= \text{Average inference latency for the selected model}, \\
\overline{P}\big(M(n_i)\big) &= \text{Average power consumption for the selected model}.
\end{align*}

\[
\text{Score} = \frac{1}{N} \sum_{i=1}^{N} \left[ \alpha\, \overline{R}\big(M(n_i)\big) - \beta\, \overline{L}\big(M(n_i)\big) - \gamma\, \overline{P}\big(M(n_i)\big) \right],
\]

\noindent where:
\begin{itemize}
    \item \( n_i \) is the thought count for the \(i\)-th prompt,
    \item \( T_1 \) and \( T_2 \) are the thresholds with \( T_1 < T_2 \),
    \item \( \alpha \), \( \beta \), and \( \gamma \) are the weights for ROUGE-L, latency, and power consumption (set to 1, 0.5, and 0.3 respectively),
    \item \( N \) is the number of prompts.
\end{itemize}

To determine the optimal threshold for routing, we conduct preliminary experiments using the previously introduced labeled MathInstruct dataset. For each threshold, we calculate an overall performance score that includes accuracy, power and latency. The objective is to find the threshold values that maximize optimize these measures.

To search the threshold space, we use a Tree-structured Parzen Estimator (TPE). TPE constructs density functions based on threshold configurations, favoring configurations that yield a higher performance score and lowering probabilities for ones that are less effective. By comparing the density functions, it determines optimal threshold values \cite{watanabe_tpe_tutorial2023}. 

\subsection{Routing With Number of Thoughts Thresholds}
The routing approach relies on deriving the optimal thresholds which determine the model used for the inference process. Our search revealed two configurations: 

\textbf{Configuration 1:}
The first search sets the threshold at 35.417 and the second at 35.418 thought steps. Essentially, this configuration bypasses the model with 7 billion parameters as prompts requiring less than 35 thought steps are handled by the 1.5B model, while those exceeding 35 by the 14B model. This high thresholds suggest that most problems can be solved by the smaller 1.5B model and only the most challenging problems should be routed to the 14B model.

\textbf{Configuration 2:}
The other configuration sets 4.80 at the first threshold and 20.26 at the second threshold. Under this setup, routing is distributed across all three models. Prompts with less than 5 thought steps are processed by the 1.7B model, those with 5 to 20 thought steps are directed to the 7B model, and prompts exceeding 20 thought steps are assigned to the 14B model.

To evaluate the effectiveness of the presented routing strategy, we compared the overall performance against the baseline where a single model is used for all prompts. As shown in Tables \ref{tab:combined_metrics} and \ref{tab:compact_metrics}, the routing approach results in improvements in latency, with only minor changes in accuracy. Thereby, reserving the largest model for the hardest prompts appears to be a viable strategy for optimizing efficiency without significantly impacting performance.

\begin{table}[ht]
\centering
\begin{tabularx}{\textwidth}{l*{6}{>{\centering\arraybackslash}X}}
\toprule
 & \multicolumn{3}{c}{\textbf{Baseline Metrics}} & \multicolumn{3}{c}{\textbf{Routing \& Comparison}} \\
\cmidrule(lr){2-4} \cmidrule(lr){5-7}
\textbf{Metric} & \textbf{1.5B} & \textbf{7B} & \textbf{14B} & \textbf{Routing} & \textbf{Absolute Improvement} & \textbf{\% Improvement} \\
\midrule
Latency (s) & 6.5125 & 7.1988 & 12.3550 & 6.5251 & 2.1636 s & 24.91\% \\
ROUGE-L   & 0.2074 & 0.2078 & 0.2056 & 0.2074 & 0.0005  & 0.22\%  \\
Power (W) & 63.9114 & 63.9114 & 63.9114 & 63.9114 & 0.0000 W & 0.00\%  \\
\bottomrule
\end{tabularx}
\caption{Baseline Metrics (Per Model), Routing Metrics, and Comparison. The routing approach uses thresholds set at 35 thoughts.}
\label{tab:combined_metrics}
\end{table}

When the routing strategy includes the 7 billion parameter model, we can still identify performance improvements. However, these improvements are smaller compared to the configuration only containing the smallest and largest model. This result shows that the number of thought metric has limitations in clearly differentiating prompts with medium difficulty level.

\begin{table}[H]
\centering
\begin{tabular}{lcccc}
\toprule
\textbf{Metric} & \textbf{Baseline Avg} & \textbf{Routing} & \textbf{Absolute Improvement} & \textbf{\% Improvement} \\
\midrule
Latency (s) & 8.7197 & 7.4154 & 1.3043 s & 14.96\% \\
ROUGE-L   & 0.2063 & 0.2083 & 0.0020   & 0.97\%  \\
Power (W) & 63.9114 & 63.9114 & 0.0000 W & 0.00\%  \\
\bottomrule
\end{tabular}
\caption{Comparison of Baseline (average over all models) and Routing Metrics. Thresholds: \(T_1=4.80697\), \(T_2=20.26065\).}
\label{tab:compact_metrics}
\end{table}

Another limitation might be introduced by the prompting strategies used to derive the number of thoughts metric. Our observations indicate differences between models. While DeepSeek models produce a more sequential and clearer reasoning step, TinyR1 tens to generate aggregated thought steps. Since the original prompt was optimized for Deepseek, it might not generalize well to other models

\section{Limitations and Threaths to Validity}\label{sec:limit}
While our work provides valuable insights into the usability of the number-of-thoughts metric, several limitations must be acknowledged. Reasoning models, such as Deepseek, require substantial computational resources to generate even a single response. Due to these resource constraints and the need to perform multiple runs per query to accurately calculate averages and standard deviations, we limited our experiments to the first 100 samples of the MathInstruct dataset. This restricted dataset size may affect the generalizability of our findings.

The mean number of thought steps was found to be 8.78, with a standard deviation of 9.6. Nevertheless, the model demonstrates strong adaptability when processing new prompts. When the trained estimator was applied to the DIA dataset, results indicated greater consistency, with mean numbers of thought steps being 7.58, 8.06, and 7.47 for easy, medium, and hard prompts, respectively. The corresponding standard deviations were 0.77, 0.64, and 1.07.

Another limitation pertains to the scope of our experiments, which involved only distilled versions of the Deepseek R1 model. While this choice provides an opportunity to explore routing scenarios across models of varying sizes, our study might not fully capture the reasoning capabilities inherent in the full-scale Deepseek R1 model.

Additionally, methodological limitations stemming from resource constraints affected our ability to comprehensively annotate the MathInstruct dataset with the number of thought steps. Availability of a larger annotated dataset would likely yield more robust results, particularly in distinguishing between medium and hard prompts. In this study, the smallest available Deepseek R1 model (1.7B parameters) was used for annotation, providing a baseline for analysis. However, this smaller model may struggle to identify nuanced differences between easy and medium prompts. Utilizing a larger model for annotation might improve differentiation between easy and medium prompts but could also lead to overgeneralization, complicating the distinction between medium and hard prompts.

Overall, although the number-of-thoughts metric shows considerable promise, further research is necessary to fully capture and differentiate the complexity associated with varying prompt difficulties.

\section{Conclusion} \label{sec:conclusion}

In this work, we utilized metadata generated from Chain-of-Thought (CoT) prompting to derive a novel metric called Number of Thoughts (NofT). This metric allowed us to effectively measure prompt complexity and provided valuable insights for various applications including adversarial prompt detection and intelligent routing of prompts to optimize inference efficiency.

\begin{itemize} 

\item \textbf{RQ1}: \textit{Can the number of reasoning steps (``number of thoughts'') generated during Chain-of-Thought prompting serve as a reliable metric for estimating task complexity?}  \\
\textbf{Answer}: The Number of Thoughts (NofT) metric was introduced as an effective quantitative measure for assessing task complexity in Chain-of-Thought (CoT) prompting. By analyzing two distinct CoT datasets, we demonstrated that NofT accurately differentiates simpler prompts from more complex ones. Furthermore, a Random Forest classifier trained on the NofT metric successfully predicted the average number of reasoning steps required by CoT models in a model agnostic manner, underscoring the metric’s reliability and predictive power in quantifying task complexity.

\item \textbf{RQ2}: \textit{Can an abnormal or unexpected number of reasoning steps effectively indicate adversarial intent, such as prompt injection attacks?}  \\
\textbf{Answer}: The abnormality in the number of thoughts generated proved to be a statistically significant indicator of adversarial intent, as confirmed by an independent t-test (t-statistic = -4.19, \(\alpha = 0.05\)). Experiments on both the Deepset and DIA-Bench datasets further demonstrated that classifiers using the NofT metric effectively detected adversarial prompts. Notably, the optimized classifier on DIA-Bench achieved 95\% accuracy with a precision of 90\% for adversarial prompts by setting a domain-informed probability threshold, thereby confirming the NofT metric's utility in security-focused prompt evaluation.

\item \textbf{RQ3}: \textit{Can the complexity estimation derived from the \"number of thoughts\" metric be effectively utilized for intelligent routing to optimize accuracy, latency, and power consumption?}  \\
\textbf{Answer}: Routing experiments demonstrated the effectiveness of leveraging the NofT metric to guide intelligent model assignment. By establishing optimal thresholds through Tree-structured Parzen Estimator (TPE) optimization, we identified configurations that significantly reduced latency (up to 24.91\%) while maintaining comparable accuracy (ROUGE-L). Specifically, simpler prompts could be effectively routed to smaller, resource-efficient models, while complex prompts were allocated to larger, more capable models. Although the NofT metric had limitations in clearly distinguishing prompts of medium complexity, the overall results substantiate its practical utility in optimizing efficiency and resource utilization in CoT-based inference pipelines.

\end{itemize}

In conclusion, the Number of Thoughts metric introduced in this study serves as a robust quantitative tool for assessing task complexity in CoT-based prompting. Its effectiveness extends beyond complexity estimation, demonstrating significant practical implications in adversarial prompt detection and resource-aware model routing strategies. These findings highlight the metric’s potential to enhance both security and operational efficiency in large language model deployments.

A Google Colab environment is leveraged in this research. A GitHub repository for this project can be found here:  
\url{https://github.com/rymarinelli/Number_Of_Thoughts/tree/main}.

%
%
%
 \bibliographystyle{splncs04}
 \bibliography{references}

\end{document}